\documentclass[10pt,twocolumn,letterpaper]{article}

\usepackage{cvpr}              
\usepackage{booktabs}
\usepackage{amssymb}
\usepackage{bbding}
\usepackage{pifont}
\usepackage{wasysym}
\usepackage{colortbl}
\usepackage[normalem]{ulem}
\useunder{\uline}{\ul}{}
\usepackage{multirow}
\definecolor{dullyellow}{RGB}{204, 204, 0}
\usepackage[dvipsnames]{xcolor}

\definecolor{cvprblue}{rgb}{0.21,0.49,0.74}
\usepackage[pagebackref,breaklinks,colorlinks,citecolor=cvprblue]{hyperref}

\def\paperID{7696} 
\def\confName{CVPR}
\def\confYear{2024}

\title{ASAM: Boosting Segment Anything Model with Adversarial Tuning}

\author{Bo Li \quad Haoke Xiao \quad Lv Tang\thanks{
Lv Tang is the corresponding author of this paper} \\ vivo Mobile Communication Co., Ltd \\ {\tt\small \{libra,xiaohaoke,lvtang\}@vivo.com}}

\begin{document}
\maketitle

\begin{abstract}
In the evolving landscape of computer vision, foundation models have emerged as pivotal tools, exhibiting exceptional adaptability to a myriad of tasks. Among these, the Segment Anything Model (SAM) by Meta AI has distinguished itself in image segmentation. However, SAM, like its counterparts, encounters limitations in specific niche applications, prompting a quest for enhancement strategies that do not compromise its inherent capabilities. This paper introduces ASAM, a novel methodology that amplifies SAM's performance through adversarial tuning. We harness the potential of natural adversarial examples, inspired by their successful implementation in natural language processing. By utilizing a stable diffusion model, we augment a subset (1\%) of the SA-1B dataset, generating adversarial instances that are more representative of natural variations rather than conventional imperceptible perturbations. Our approach maintains the photorealism of adversarial examples and ensures alignment with original mask annotations, thereby preserving the integrity of the segmentation task. The fine-tuned ASAM demonstrates significant improvements across a diverse range of segmentation tasks without necessitating additional data or architectural modifications. The results of our extensive evaluations confirm that ASAM establishes new benchmarks in segmentation tasks, thereby contributing to the advancement of foundational models in computer vision. Our project page is in \url{https://asam2024.github.io/}.
\end{abstract}

\section{Introduction}
The concept of foundation models has been pivotal in advancing the fields of natural language processing (NLP) and, more recently, computer vision. Originating in NLP with influential models such as BERT~\cite{DBLP:conf/naacl/DevlinCLT19}, the GPT series~\cite{DBLP:journals/corr/abs-2303-08774}, LLaMA~\cite{DBLP:journals/corr/abs-2302-13971} and PaLM~\cite{DBLP:journals/jmlr/ChowdheryNDBMRBCSGSSTMRBTSPRDHPBAI23}, these models have showcased remarkable zero-shot generalization capabilities to unseen tasks. This success has spurred the development of similar paradigm-shifting models in computer vision. These visual foundation models, such as DINOv2~\cite{DBLP:journals/corr/abs-2304-07193}, CLIP~\cite{DBLP:conf/icml/RadfordKHRGASAM21}, BLIP~\citep{DBLP:conf/icml/0001LXH22}, SAM~\cite{Kirillov_2023_ICCV} and Stable Diffusion~\cite{DBLP:conf/cvpr/RombachBLEO22}, demonstrate remarkable zero-shot capabilities and broad generalization across various tasks.

\begin{figure}
    \centering
    \includegraphics[width=\linewidth]{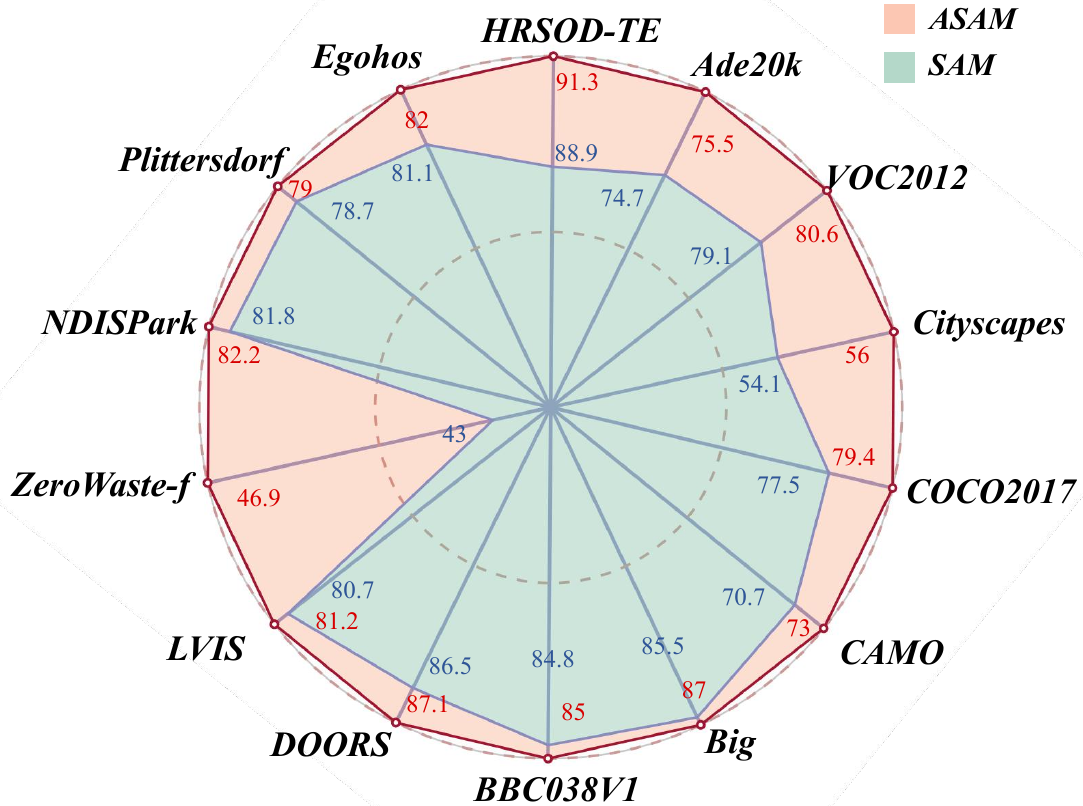}
    \caption{Performance comparison between ASAM and SAM on diverse segmentation datasets across different downstream tasks.}
    \vspace{-0.5cm}
    \label{fig:radar}
\end{figure}

Among them, Segment Anything Model (SAM) stands out as a pioneering  visual foundation model specializing in image segmentation. Trained on over 1 billion masks from a massive visual corpus, SAM has revolutionized the field with its ability to segment a diverse range of objects and structures across various scenarios. Despite its impressive performance, SAM, like any foundational model, has areas where it can be further enhanced~\cite{DBLP:journals/corr/abs-2304-04709,DBLP:journals/corr/abs-2304-05750,DBLP:journals/corr/abs-2304-06022}.

An important research direction is identifying SAM's limitations on certain  downstream tasks and developing techniques to boost its performance. Many techniques have explored like fine-tuning~\cite{DBLP:journals/corr/abs-2304-12306,DBLP:journals/corr/abs-2304-13785,DBLP:journals/corr/abs-2306-12737,DBLP:journals/corr/abs-2308-14604} and adapter modules~\cite{DBLP:journals/corr/abs-2304-09148,DBLP:journals/corr/abs-2401-02326,DBLP:journals/corr/abs-2304-12620} to specialize the SAM for specific downstream tasks. While fine-tuning unlocks the potential of SAM for a specific task, it compromises the model’s inherent generalization capabilities~\cite{DBLP:journals/corr/abs-2306-01567}. Alternative approaches preserve SAM's original parameters, adding adaptation layers or post-processing modules~\cite{DBLP:journals/corr/abs-2306-01567,DBLP:journals/corr/abs-2307-04767}. Those methods, though effective, require additional parameters and annotated training data, limiting its scalability and efficiency.


The above challenges bring us to the core motivation of this work: How can we further boost the generalization ability of SAM as a foundational vision model without relying on substantial extra data, altering its base architecture, or compromising its zero-shot capabilities? So that we can unlock SAM's potential while keeping its broad applicability across vision tasks. Existing solutions, while effective in specific contexts, do not address the fundamental challenge of enhancing SAM's inherent performance across a diverse range of scenarios.

In response to this challenge, we turn to the realm of NLP for inspiration, particularly its pioneering advancements in foundational model research. The unique successes~\cite{DBLP:conf/iclr/ZhuCGSGL20} observed in adversarial training (AT) within NLP offer a new vantage point. In contrast to the visual domain, where standard adversarial training often necessitates a compromise between robustness and model performance~\cite{DBLP:conf/icml/WenLJ20}, AT in NLP not only strengthens model robustness but also concurrently bolsters generalization and accuracy~\cite{DBLP:conf/iclr/ZhuCGSGL20}. This divergence is believed to be attributed to the closer resemblance of adversarial examples in natural language to real-world textual scenarios, such as common human spelling errors. We surmise that the triumph of adversarial training in NLP is derived from the ``realness" and ``naturalness" of its generated adversarial examples. This insight leads us to explore the possibility of adapting adversarial training techniques, which have been successful in NLP, to visual foundation models like SAM. This approach aims to apply cross-disciplinary insights innovatively to improve specific tasks in computer vision.

Applying the above concept to SAM, our approach aims to utilize ``natural" adversarial examples akin to those found in NLP to elevate visual foundation models. Inspired by the effective tuning methodologies in NLP~\cite{DBLP:journals/corr/abs-2204-07705,DBLP:conf/nips/Ouyang0JAWMZASR22,DBLP:conf/acl/WangKMLSKH23}, we propose fine-tuning the SAM using these more ``natural" adversarial examples, thereby circumventing the high costs often associated with conventional adversarial training. Traditional methods for generating visual adversarial examples typically adhere to $l_p$ norm constraints, resulting in perturbations that are not entirely natural and exhibit a domain shift from real-world noise. This leads to a disparity between such adversarial examples and the genuinely challenging examples encountered in real-world scenarios~\cite{dra}.

To generate adversarial examples that are both natural and photorealistic for tuning SAM, we are inspired by recent adversarial attack~\cite{DBLP:conf/nips/ChenLWJDZ23} and hypothesize that natural images can be projected onto a low-dimensional manifold via a generative model~\cite{DBLP:conf/cvpr/RombachBLEO22}. This manifold, trained on natural images, ensures the photorealism and richness of content. By mapping an image onto this manifold and then shifting it along an adversarial direction within the manifold, we can produce adversarial examples that are both natural and photorealistic. To maintain the consistency of object shapes with the original mask labels during the back-mapping process, we incorporate an additional mask prompt branch in the generative model. This integration ensures that the adversarial examples are not only realistically aligned but also accurately correspond to their original mask labels. Ultimately, by fine-tuning a select subset of parameters in a large vision model with these naturally realistic and accurately aligned adversarial examples, we achieve significant enhancements in performance. In conclusion, our work makes several key contributions:

\begin{itemize}

\item Drawing inspiration from the successes in NLP, we introduce a novel framework, termed adversarial tuning, aimed at enhancing the generalization abilities of visual foundation models like SAM. This approach represents an innovative application of cross-disciplinary insights to address specific challenges in computer vision tasks. 

\item By projecting natural images onto a low-dimensional manifold using a generative model, we generate adversarial examples that are both natural and photorealistic. We further enhance this approach by integrating a mask prompt branch into the generative model, ensuring that the adversarial examples maintain consistency with the original mask labels in terms of object shape. 
 
\item Leveraging our approach, we fine-tune SAM with ``natural" adversarial examples, derived from just 1\% of the SA-1B dataset, resulting in an enhanced version termed ASAM. To validate ASAM's effectiveness, we conduct extensive quantitative and qualitative analyses. As shown in Fig. \ref{fig:radar}, ASAM has achieved significant improvements in SAM's performance across a wide range of segmentation datasets and various downstream tasks.

\end{itemize}

\section{Related Works}
\subsection{Segment Anything Model (SAM)}
Meta Research team has released the ``Segment Anything" project~\cite{Kirillov_2023_ICCV}. This project develops the SAM and an extensive dataset, SA-1B, featuring over 1 billion masks on 11 million licensed and privacy-respecting images. Designed for prompt-based segmentation, SAM is capable of zero-shot adaptation to new image distributions and tasks. As a pioneering  visual foundation model, its zero-shot segmentation abilities and prompt-based approach have facilitated rapid application in diverse areas, going beyond image segmentation to tasks like 3D understanding and video processing ~\cite{DBLP:journals/corr/abs-2305-05803,DBLP:journals/corr/abs-2305-01443,DBLP:journals/corr/abs-2305-13620,DBLP:journals/corr/abs-2306-04121,DBLP:journals/corr/abs-2305-12659,DBLP:journals/corr/abs-2306-03908,DBLP:journals/corr/abs-2309-05499,DBLP:journals/corr/abs-2310-10912}.

While SAM's capability is impressive, its effectiveness in real-world scenarios, such as medical images and other challenging segmentation conditions, has been a topic of investigation. Difficulties arise when {segmenting minuscule and slender objects~\cite{DBLP:journals/corr/abs-2306-01567}, objects with obscure boundaries~\cite{DBLP:journals/corr/abs-2305-11513,DBLP:journals/corr/abs-2304-06022}}, camouflaged objects~\cite{DBLP:journals/corr/abs-2304-04709,DBLP:journals/corr/abs-2304-05750,DBLP:journals/corr/abs-2304-06022}, and transparent objects~\cite{DBLP:journals/corr/abs-2305-00278}. Just like any foundational model, SAM has areas where it can be further enhanced.

To address these challenges, researchers have introduced various methods. For instance, the work~\cite{DBLP:journals/corr/abs-2304-12306} proposes a straightforward fine-tuning approach to tailor the SAM for general medical image segmentation. Rigorous experimentation on both 3D and 2D segmentation tasks illustrates that MedSAM surpasses the default SAM. SAM-Adapter~\cite{DBLP:journals/corr/abs-2304-09148, DBLP:journals/corr/abs-2304-12620} leverages domain specific information or visual prompts to enhance the segmentation network through the use of simple yet effective adapters. By combining task-specific knowledge with general knowledge learned by the large model, SAM-Adapter can notably improve the performance of SAM in challenging tasks. While fine-tuning unlocks the potential of SAM for a specific task, it compromises the model’s inherent generalization capabilities~\cite{DBLP:journals/corr/abs-2306-01567}.  Alternative approaches preserve SAM's original parameters, adding adaptation layers or post-processing modules like in SAM-HQ~\cite{DBLP:journals/corr/abs-2306-01567} and Semantic-SAM~\cite{DBLP:journals/corr/abs-2307-04767}. Those methods, though effective, require additional parameters and annotated training data, limiting its scalability and efficiency.
Additionally, instead of direct modifying SAM's parameters, refining the input prompt~\cite{DBLP:journals/corr/abs-2305-03048} or output of SAM~\cite{DBLP:journals/corr/abs-2304-07764,DBLP:journals/corr/abs-2305-09418} are also viable strategies.

Our approach diverges from these existing methods, aiming to further enhance SAM's generalization capabilities as a foundational vision model. We seek to achieve this without substantial reliance on extra data, alterations to its architecture, or compromising its zero-shot capabilities.

\subsection{Adversarial Examples \& Adversarial Training}
Adversarial examples, in computer vision, are deliberately modified inputs designed to cause misclassification by a model~\cite{DBLP:journals/corr/SzegedyZSBEGF13,DBLP:journals/corr/GoodfellowSS14}. These perturbations, initially defined as imperceptible variations in image pixels within small $l_1$, $l_2$, and $l_{\infty}$ norms (uniformly referred as $l_p$), form the basis for understanding adversarial vulnerabilities in visual models. AT, proposed as an effective defense mechanism, aims to enhance robustness by training models with these adversarial examples~\cite{DBLP:conf/iclr/MadryMSTV18}. However, it has been observed that AT often leads to a trade-off between adversarial robustness and clean accuracy, presenting a challenge to model generalization~\cite{DBLP:conf/icml/ZhangYJXGJ19,DBLP:conf/iclr/TsiprasSETM19}.  Despite great efforts~\cite{DBLP:conf/icml/RaghunathanXYDL20,DBLP:conf/iclr/RadeM22,DBLP:journals/corr/abs-2012-13628} have been made for mitigating this trade-off, the bad generalization of AT still cannot be fully remedied till now. 

In contrast, the NLP realm exhibits a different trend: AT has been found to enhance both the generalization and robustness of language models~\cite{DBLP:conf/iclr/MiyatoDG17,DBLP:journals/pami/MiyatoMKI19,DBLP:conf/acl/ChengJM19}. Recent studies like the work~\cite{DBLP:conf/iclr/ZhuCGSGL20} demonstrate that AT can even boost the performance of transformer-based language foundational models. The work~\cite{DBLP:conf/nips/MaoCDZQYLZ022} wants to directly copy the success of AT in NLP to enhance the visual features, suggesting discrete representation as a key factor. Although they generate adversarial examples with more imperceptible perturbations than traditional $l_p$ perturbations, the perturbations are still not entirely natural and exhibit domain shift from real-world noise. In this paper, we surmise that the triumph of AT in NLP is derived from the ``realness" and ``naturalness" of its adversarial examples. 

Notably, there have been attempts to use AT for improving clean accuracy in vision tasks. The work~\cite{DBLP:conf/cvpr/XieTGWYL20} employs split batch norms to separate clean and adversarial example statistics, enhancing adversarial feature learning for generalization. However, this operation is not applicable to transformer-based modern foundation models~\cite{DBLP:conf/naacl/DevlinCLT19,DBLP:journals/corr/abs-2303-08774,Kirillov_2023_ICCV}. Another related work to ours is ~\cite{DBLP:journals/corr/abs-2012-13628}, which although similar in name, focuses on using fine-tuning to replace adversarial training to obtain adversarial robustness at low cost. Inspired by works~\cite{DBLP:conf/cvpr/RombachBLEO22,DBLP:conf/nips/ChenLWJDZ23} and NLP, we introduce a novel framework ASAM, fine-tuning SAM with ``natural" adversarial examples. This approach paves a new path for enhancing visual foundation models, leveraging the ``realness" and ``naturalness" of adversarial examples to augment SAM's generalization capabilities without substantial additional data or major architectural changes.

\section{Method}

\begin{figure*}
    \centering
    \includegraphics[width=0.92\linewidth]{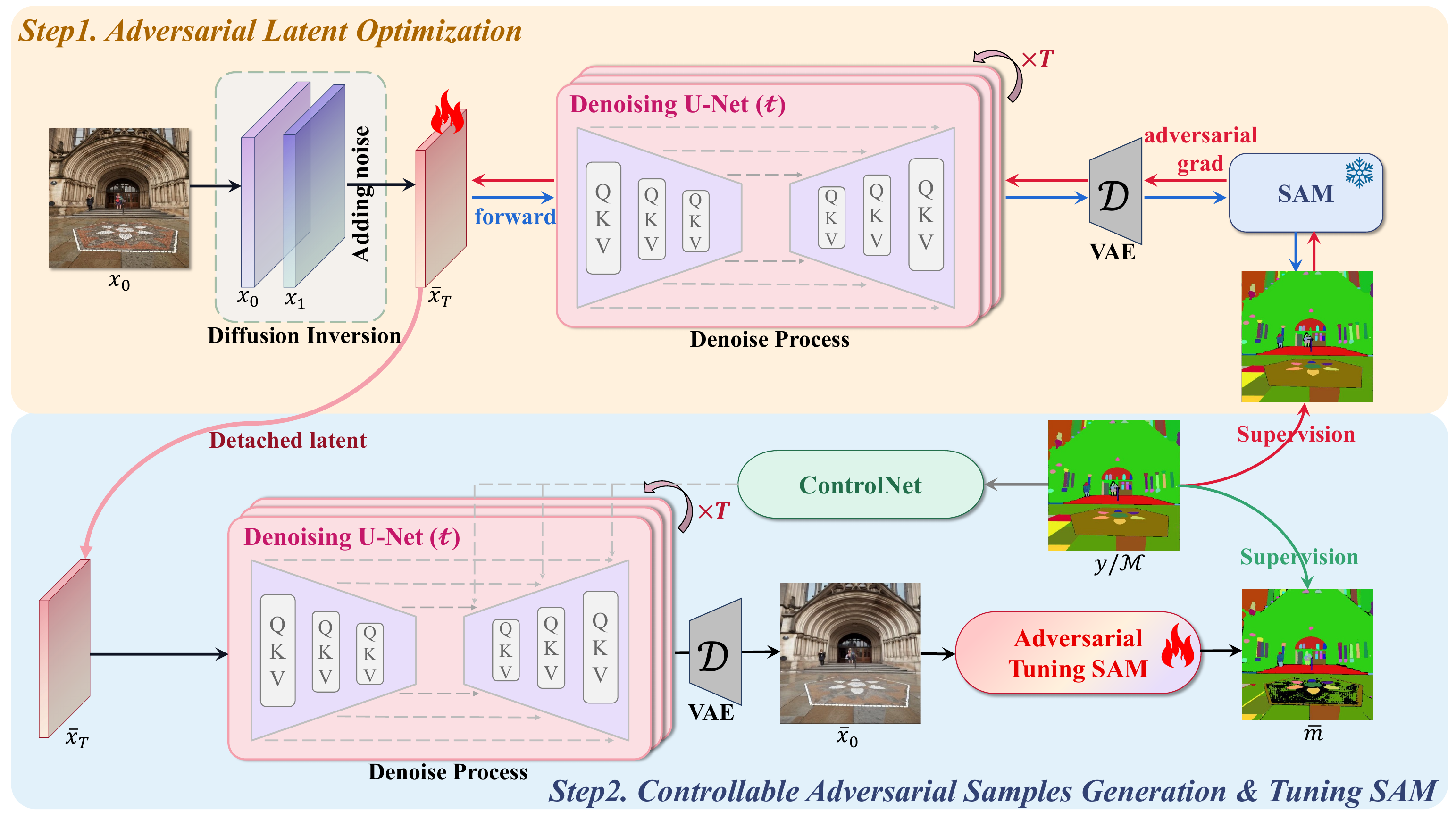}
    \caption{The architecture of our proposed ASAM framework. In the first step, we project the input image into the latent space and then optimize the latent space with adversarial technologies. In the second step, we use the optimized latent to generate adversarial samples controlled by masks. Finally, we fine-tune the SAM with the generated ``natural" adversarial samples.}
    \vspace{-0.5cm}
    \label{fig:pipline} 
\end{figure*}

\subsection{Overview}
We aim to generate ``natural" adversarial images from the SA-1B \cite{Kirillov_2023_ICCV} dataset, and subsequently, employ these generated images along with corresponding SA-1B masks to fine-tune SAM. Note that, during fine-tuning the SAM, we do not modify the SAM structure and incorporate any extra annotated data. Therefore, our proposed ASAM framework achieves the goal of enhancing the generalizability of SAM solely based on its inherent data and structural characteristics. Our proposed ASAM framework mainly contains two steps which are described in detail in the following. 

\noindent \textbf{Adversarial Latent Optimization.} Existing methods~\cite{DBLP:conf/iclr/ZhuCGSGL20,DBLP:conf/cvpr/JiaZW0WC22,DBLP:conf/icml/RiceWK20,DBLP:conf/icml/ZhangXH0CSK20,DBLP:conf/iclr/ZhangZ00SK21} for generating adversarial images typically adhere to $l_p$ norm constraints, resulting
in perturbations that are not entirely natural and exhibit a domain shift from real-world noise. In this paper, to generate adversarial examples that are both natural and photorealistic for tuning SAM, we hypothesize that natural images can be first projected onto a low-dimensional manifold via a generative model, such as Stable Diffusion \cite{DBLP:conf/cvpr/RombachBLEO22}. Subsequently, by optimizing the low-dimensional manifold, we are able to search for a suitable adversarial latent representation, allowing for a re-projection into the natural image domain effectively. We illustrate the process of optimizing adversarial latent representation in \textit{Sec. \ref{method:step1}}.

\noindent \textbf{Controllable Adversarial Samples Generation.} The above optimization process adds slight perturbations to the latent representation. Therefore, the naive re-projection may result in the generated adversarial images not aligning properly with the corresponding SA-1B masks. To address this issue, after the optimization is completed, we further design the control branch, which leverages the ControlNet \cite{DBLP:journals/corr/abs-2302-05543} to guide the re-projection process. More details about this process are described in \textit{Sec. \ref{method:step2}}.

\subsection{Adversarial Latent Optimization} \label{method:step1}
Herein, we demonstrate the methodology for searching the adversarial latent representation of SA-1B images within the low-dimensional manifold space of the generative model. Taking into account the balance between computational expense and images quality, we opt for Stable Diffusion as our generative model to produce low-dimensional latent representations. Subsequently, we optimize the generated latent representation which enables the creation of diverse adversarial images. 
 
\subsubsection{Projecting Image to Diffusion Latent} \label{step1.1}
The diffusion inversion is commonly used for projecting the image to low-dimensional latent space. In the case of the diffusion model, we employ the DDIM inversion technique \cite{DBLP:conf/iclr/SongME21} which utilizes the conditional embedding $C = \psi(P)$ derived from prompts $P$ using CLIP text encoder, predicated on the premise that the ordinary differential equation procedure is reversible within a finite number of steps:
\begin{align}
    x_{t+1} = \sqrt{\frac{\alpha_{t+1}}{\alpha_{t}}}x_t + (\sqrt{\frac{1}{\alpha_{t+1}}-1} - \sqrt{\frac{1}{\alpha_t-1}}) \cdot \epsilon_\theta(x_t,t,C).
\end{align}
Given an image $x_0$, we use a schedule $\{\beta_1, \ldots, \beta_T\} \in (0,1)$, with $\alpha_t = \prod_{i=1}^{t} (1-\beta_i)$ following \cite{DBLP:conf/iclr/SongME21}. This approach effectively operates in the opposite direction to the denoising process (i.e., $x_0 \rightarrow x_T$ rather than $x_T \rightarrow x_0$), projecting the image $x_0$ into the latent space at $x_T$. The text description of each image is generated through BLIPv2~\cite{DBLP:conf/icml/0008LSH23}.

Text-to-image synthesis frequently emphasizes the role of the prompt, culminating in the introduction of a classifier-free guidance approach~\cite{DBLP:journals/corr/abs-2207-12598}. This method generates predictions with no condition and merges them with predictions that are conditioned on specific inputs. Let $\omega$ represents the guidance scale factor and $\emptyset = \psi(`` ")$ denotes the embedding for an empty text prompt, then the formula for classifier-free guidance is articulated as follows:
\begin{align} 
\Tilde{\epsilon_\theta}(x_t,t,C,\emptyset) = \omega \cdot \epsilon_\theta(x_t,t,C) + (1-\omega) \cdot \epsilon_\theta(x_t,t,\emptyset).
\end{align} 
$\omega = 7.5$ is adopted as the standard setting for Stable Diffusion. During the reverse process of DDIM sampling, the model $\epsilon_\theta$ forecasts the noise, which might introduce minor inaccuracies at each step. Given its substantial guidance scale parameter $\omega$, the classifier-free guidance method is prone to magnifying these small errors, resulting in a build-up of inaccuracies. Thus, utilizing the reverse DDIM sampling process alongside classifier-free guidance not only disrupts the Gaussian noise distribution but also generates visual anomalies that compromise realism~\cite{DBLP:conf/cvpr/MokadyHAPC23}.


To mitigate the accumulation of errors, our approach is inspired by the strategy outlined in~\cite{DBLP:conf/cvpr/MokadyHAPC23}, where we optimize a distinct null text embedding $\emptyset_t$ for each timestep $t$. Initially, executing the DDIM inverse sampling process with $\omega = 1$ yields a series of successive latent representations $\{x_0^*, ..., x_T^*\}$, starting with $x_0^* = x_0$. Subsequently, we embark on an optimization process for the timesteps $t = \{T, ..., 1\}$, employing $\omega = 7.5$ and setting $\bar{x}_T = x_T^*$:
\begin{align}
\underset{\emptyset_t}{min} || x_{t-1}^* - x_{t-1}(\bar{x}_t,t,C,\emptyset_t) ||_2^2.
\end{align}
For ease of understanding, let $x_{t-1}(\bar{x}_t,t,C,\emptyset_t)$ denote the DDIM sampling step, where $\bar{x}_{t}$ serves as the input latent, $\emptyset_t$ as the null text embedding, and $C$ is the text embedding. Upon finishing each step, $\bar{x}_{t-1}$ is updated in accordance with the equation:

\begin{align}
\bar{x}_{t-1} = x_{t-1}(\bar{x}_t,t,C,\emptyset_t).
\end{align}

Finally, we can achieve the latent representation $\bar{x}_T = x_{T}^*$ with the optimized null text embedding $\{\emptyset_t\}_1^T$ generated by the diffusion model.
We exploit this latent in the low-dimensional manifold to generate adversarial images.

\subsubsection{Adversarial Optimization of Latent}
\label{step1.2}
In this section, we undertake an optimization of the latent representation to enhance the generation of natural adversarial images. Within the latent space established by Sec. \ref{step1.1}, the null text embedding $\emptyset_t$ ensures the quality of the reconstructed image, whereas the text embedding $C$ retains the semantic content of the image. Consequently, optimizing both embeddings simultaneously may not lead to optimal outcomes. Considering that the noise $\bar{x}_T$ significantly encapsulates the image's details in the latent space, we opt to focus our optimization efforts on it.

Building upon the latent representation generated in Sec. \ref{step1.1}, we characterize the denoising procedure of the diffusion model as $\Omega(\cdot)$, implemented via the DDIM sampling step. This process encompasses $T$ iterations:
\begin{align}
    &\Omega(x_t,T,C,\left\{\emptyset_t \right\}_1^T) = \\
    &x_0(x_1(...,(x_{T}, T,C,\emptyset_{T}),...,1,C,\emptyset_1),0,C,\emptyset_0). \nonumber
\end{align}
Here, $x_t$ denotes the latent variable at iteration $t$, with $T$ being the total number of iterations, $C$ standing for the additional conditioning variables, and $\{\emptyset_t\}_{1}^{T}$ signifying the sequence of null text embeddings applied at each iteration. The process concludes with the reconstructed image, represented by $\bar{x}_0 = \Omega(\bar{x}_T, T, C, \{\emptyset_t\}_1^T)$. The operations of the Variational Autoencoder (VAE) are not elaborated upon in this manuscript, given its differentiable nature. We frame our adversarial objective optimization as follows:
\begin{align}
\underset{\delta}{\max} \ \mathcal{L}(\mathcal{S}_{\theta}(\bar{x}_0),y), \ \text{s.t.} \ ||\delta||_{\infty} \leq \kappa,
\label{ddim process}
\end{align}
In this equation, $\delta$ signifies the adversarial perturbation within the latent space, $y$ represents the mask label obtained from the SA-1B dataset, and $\mathcal{S}_\theta$ denotes the SAM with a fixed parameter set $\theta$. The loss function, $\mathcal{L}$, is an amalgamation of mean square error, binary cross-entropy loss, and dice loss, articulated as $\mathcal{L} = \mathcal{L}_{mse} + \mathcal{L}_{bce} + \mathcal{L}_{dice}$. To preserve the consistency between the original image $x_0$ and its reconstructed counterpart $\bar{x}_0$, we posit that the perturbation $\delta$ exerts a minimal impact on this consistency, provided its magnitude is exceedingly slight, namely $||\delta||_{\infty} \leq \kappa$. The principal challenge is to pinpoint the optimal $\delta$ that escalates the segmentation loss. Echoing the approach of traditional adversarial strategies, we utilize gradient-based methods to approximate $\delta$ with the formula: $\delta \approx \eta \nabla_{x_{T}} \mathcal{L}(\mathcal{S}_\theta(\bar{x}_0), y)$, where $\eta$ is the scale of perturbations aligned with the gradient's direction. By applying the chain rule to unfold $\nabla_{\bar{x}_T} \mathcal{L}(\mathcal{S}_\theta(\bar{x}_0),y)$, we delineate each derivative component:
\begin{align}
\nabla_{\bar{x}_T}\mathcal{L}(\mathcal{S}_\theta(\bar{x}_T),y) = \frac{\partial \mathcal{L}}{\partial \bar{x}_0} \cdot \frac{\partial \bar{x}_0}{\partial \bar{x}_1} \cdot \frac{\partial \bar{x}_1}{\partial \bar{x}_2} \cdots \frac{\partial \bar{x}_{T-1}}{\partial \bar{x}_T}.
\end{align}

\subsection{Controllable Adversarial Samples Generation} 
\label{method:step2}
After obtaining an adversarial latent representation, a reverse diffusion process can be employed to generate the final adversarial examples. However, the optimization process in Stable Diffusion space would introduce minor disturbances to the adversarial latent variables, which would result in misalignment between the generated image's shape and its corresponding label. Intuitively, this issue could potentially be addressed by using more precise prompts in the diffusion model. Nonetheless, the capability of text prompts to control the spatial shape of images is limited, as it's challenging to describe the exact shape of objects through text alone. To overcome this limitation, we additionally train a mask-to-image ControlNet inserted into the reverse process, which offers enhanced spatial shaping capabilities.

\begin{table*}[]
\centering
\caption{Zero-shot segmentation result mIoU comparison on 14 datasets using box prompt.}
\scalebox{0.78}{
\begin{tabular}{@{}c|cccccccc@{}}
\toprule
\cellcolor[HTML]{FFFFFF}Methods &
  \cellcolor[HTML]{FFFFFF}DOORS\cite{DBLP:journals/corr/abs-2210-16253} &
  \cellcolor[HTML]{FFFFFF}LVIS\cite{DBLP:conf/cvpr/GuptaDG19} &
  \cellcolor[HTML]{FFFFFF}ZeroWaste-f\cite{DBLP:conf/cvpr/BashkirovaAZAAH22} &
  \cellcolor[HTML]{FFFFFF}NDISPark\cite{DBLP:conf/visapp/CiampiSCGA21} &
  Egohos\cite{DBLP:conf/eccv/ZhangZSS22} &
  Plittersdorf\cite{DBLP:journals/sensors/HauckeKS22} &
  \cellcolor[HTML]{FFFFFF}BBC038V1\cite{caicedo2019nucleus} &
  Average \\ \midrule
SAM &
  86.5 &
  80.7 &
  43.0 &
  81.8 &
  81.1 &
  78.7 &
  84.8 &
  76.7 \\
\cellcolor[HTML]{FFFFFF}SAM + DAT Tuning &
  \cellcolor[HTML]{FFFFFF}65.0 &
  \cellcolor[HTML]{FFFFFF}48.1 &
  \cellcolor[HTML]{FFFFFF}35.2 &
  \cellcolor[HTML]{FFFFFF}53.2 &
  45.7 &
  31.4 &
  \cellcolor[HTML]{FFFFFF}54.6 &
  47.6 \\
\cellcolor[HTML]{FFFFFF}SAM + PGD Tuning &
  \cellcolor[HTML]{FFFFFF}68.3 &
  \cellcolor[HTML]{FFFFFF}52.5 &
  \cellcolor[HTML]{FFFFFF}38.9 &
  \cellcolor[HTML]{FFFFFF}60.0 &
  51.8 &
  40.3 &
  \cellcolor[HTML]{FFFFFF}65.8 &
  53.9 \\
\cellcolor[HTML]{FFFFFF}SAM + DatasetDM &
  \cellcolor[HTML]{FFFFFF}86.0 &
  \cellcolor[HTML]{FFFFFF}62.2 &
  \cellcolor[HTML]{FFFFFF}29.2 &
  \cellcolor[HTML]{FFFFFF}69.5 &
  53.5 &
  70.3 &
  \cellcolor[HTML]{FFFFFF}84.2 &
  65.0 \\
ASAM &
  \textbf{87.1} &
  \textbf{81.2} &
  \textbf{46.9} &
  \textbf{82.2} &
  \textbf{82.0} &
  \textbf{79.0} &
  \textbf{85.0} &
  \textbf{77.6} \\ \midrule \midrule
\cellcolor[HTML]{FFFFFF}Methods &
  \cellcolor[HTML]{FFFFFF}Ade20k\cite{DBLP:journals/ijcv/ZhouZPXFBT19} &
  \cellcolor[HTML]{FFFFFF}VOC2012\cite{everingham2012pascal} &
  \cellcolor[HTML]{FFFFFF}Cityscapes\cite{DBLP:conf/cvpr/CordtsORREBFRS16} &
  \cellcolor[HTML]{FFFFFF}COCO2017\cite{lin2014microsoft} &
  \cellcolor[HTML]{FFFFFF}HRSOD-TE\cite{DBLP:conf/iccv/ZengZLZL19} &
  \cellcolor[HTML]{FFFFFF}CAMO\cite{DBLP:journals/cviu/LeNNTS19} &
  \cellcolor[HTML]{FFFFFF}Big\cite{DBLP:conf/cvpr/ChengCTT20} &
  Average \\ \midrule
SAM &
  74.7 &
  79.1 &
  54.1 &
  77.5 &
  88.9 &
  70.7 &
  85.5 &
  75.8 \\
\cellcolor[HTML]{FFFFFF}SAM + DAT Tuning &
  \cellcolor[HTML]{FFFFFF}54.6 &
  \cellcolor[HTML]{FFFFFF}53.3 &
  \cellcolor[HTML]{FFFFFF}31.2 &
  \cellcolor[HTML]{FFFFFF}53.7 &
  \cellcolor[HTML]{FFFFFF}55.0 &
  \cellcolor[HTML]{FFFFFF}{\color[HTML]{333333} 52.5} &
  \cellcolor[HTML]{FFFFFF}{\color[HTML]{333333} 57.6} &
  51.1 \\
\cellcolor[HTML]{FFFFFF}SAM + PGD Tuning &
  \cellcolor[HTML]{FFFFFF}58.7 &
  \cellcolor[HTML]{FFFFFF}60.3 &
  \cellcolor[HTML]{FFFFFF}33.5 &
  \cellcolor[HTML]{FFFFFF}58.8 &
  \cellcolor[HTML]{FFFFFF}63.4 &
  \cellcolor[HTML]{FFFFFF}{\color[HTML]{333333} 55.1} &
  \cellcolor[HTML]{FFFFFF}{\color[HTML]{333333} 63.9} &
  56.2 \\
\cellcolor[HTML]{FFFFFF}SAM + DatasetDM &
  \cellcolor[HTML]{FFFFFF}57.4 &
  \cellcolor[HTML]{FFFFFF}43.7 &
  \cellcolor[HTML]{FFFFFF}44.9 &
  \cellcolor[HTML]{FFFFFF}54.1 &
  \cellcolor[HTML]{FFFFFF}45.4 &
  \cellcolor[HTML]{FFFFFF}26.0 &
  \cellcolor[HTML]{FFFFFF}36.4 &
  44.0 \\
ASAM &
  \textbf{75.5} &
  \textbf{80.6} &
  \textbf{56.0} &
  \textbf{79.4} &
  \textbf{91.3} &
  \textbf{73.0} &
  \textbf{87.0} &
  \textbf{77.5} \\ \toprule
\end{tabular}}
\label{totaltable1}
\vspace{-0.5cm}
\end{table*}

ControlNet adjusts the task-specific conditions within the denoising U-Net architecture, aiming to steer the overall behavior of the diffusion model more precisely. The core architecture of the Stable Diffusion model is a U-Net, consisting of an encoder, a middle block, and a decoder that utilizes skip connections. Both the encoder and decoder feature 12 blocks each, culminating in a total of 25 blocks when including the middle block. ControlNet is employed to generate a trainable duplicate of the 12 encoder blocks and the single middle block from the Stable Diffusion model. These 12 blocks are distributed across four different resolutions (64 × 64, 32 × 32, 16 × 16, 8 × 8), with each resolution comprising three blocks. The generated outputs from these blocks are then integrated into the 12 skip connections and the middle block of the Diffusion U-Net, enhancing its capability to manipulate image characteristics with greater finesse. The operation of ControlNet is denoted as $Z(\cdot;\cdot)$, and it allows for a reconfiguration of the denoising U-Net:
\begin{align}
    n = Dec(Enc(x_t,T,C,\emptyset_t),Z(x_t,T,\mathcal{M},C, \emptyset_t)),
\end{align}
where $\mathcal{M}$ is the mask prompt. 
Based on the denoising U-Net, we represent the adversarial examples reconstruction:
\begin{flalign}
    &\Omega(\bar{x_t},T, \mathcal{M},C,\left\{\emptyset_t \right\}_1^T) = \\
    &x_0(x_1(...,(\bar{x_{T}},T,\mathcal{M},C,\emptyset_{T}),...,1,\mathcal{M},C,\emptyset_1),0,\mathcal{M},C,\emptyset_0). \nonumber
\end{flalign}

\subsection{Fine-tuning SAM with Adversarial Samples} \label{method:step3}
Different from previous methods \cite{DBLP:journals/corr/abs-2306-01567,DBLP:journals/corr/abs-2304-12620,DBLP:journals/corr/abs-2304-09148,DBLP:journals/corr/abs-2307-04767} which alter the structure of SAM, our aim is to enhance the overall capabilities of the SAM without any structural modifications. The selection of appropriate parameters for fine-tuning necessitates careful consideration, taking into account factors such as efficiency and the risk of over-fitting. In this regard, we specifically choose to fine-tune the output tokens and mask token of SAM, which accounts for only approximately $0.001\%$ of the total parameters in the SAM. Additionally, to ensure fast convergence on adversarial samples while maintaining generalization, we adopt the learning rate schedule strategy ``slow start fast decay", as described in the work~\cite{DBLP:journals/corr/abs-2012-13628}. Furthermore, our proposed ASAM indicates that employing only $1\%$ samples from the SA-1B dataset already yields significant performance improvements.

\section{Experiment}
\subsection{Experimental Setting}
\textbf{Implementation Details.} We use stable-diffusion-v1-5 \cite{DBLP:conf/cvpr/RombachBLEO22} pre-trained on the LAION5B \cite{DBLP:conf/nips/SchuhmannBVGWCC22} dataset. The description of each training image is automatically generated using BLIPv2~\cite{DBLP:conf/icml/0008LSH23}. We use ControlNet v1.0 to control the generation process. We use SAM with vit-base backbone. The training dataset used in this paper is \textit{sa\_000000} subset from SA-1B dataset. For the adversarial example generation process, we set DDIM steps $T$ to 50, the number of optimization steps of null text embedding to 10, the number of attacks on adversarial samples to 10, and the attacks size $\kappa$ to 0.02. We fine-tune the SAM with 10 epochs using Adam optimizer. The learning rate first increases linearly from 0.01 to 0.05, then decay exponentially. We adopt 8 NVIDIA 48G A6000 GPUs for training.

\noindent \textbf{Evaluation Datasets.}
Following SAM \cite{Kirillov_2023_ICCV}, we evaluate ASAM on datasets and tasks that are not seen during training. The evaluation datasets may include novel image distributions, such as underwater or ego-centric images that, to our knowledge, do not appear in SA-1B. We use a newly compiled suite of 14 datasets with diverse image distributions under mIoU evaluation, as shown in Table. \ref{totaltable1}.

\begin{figure*}
    \centering
    \includegraphics[width=0.85\linewidth]{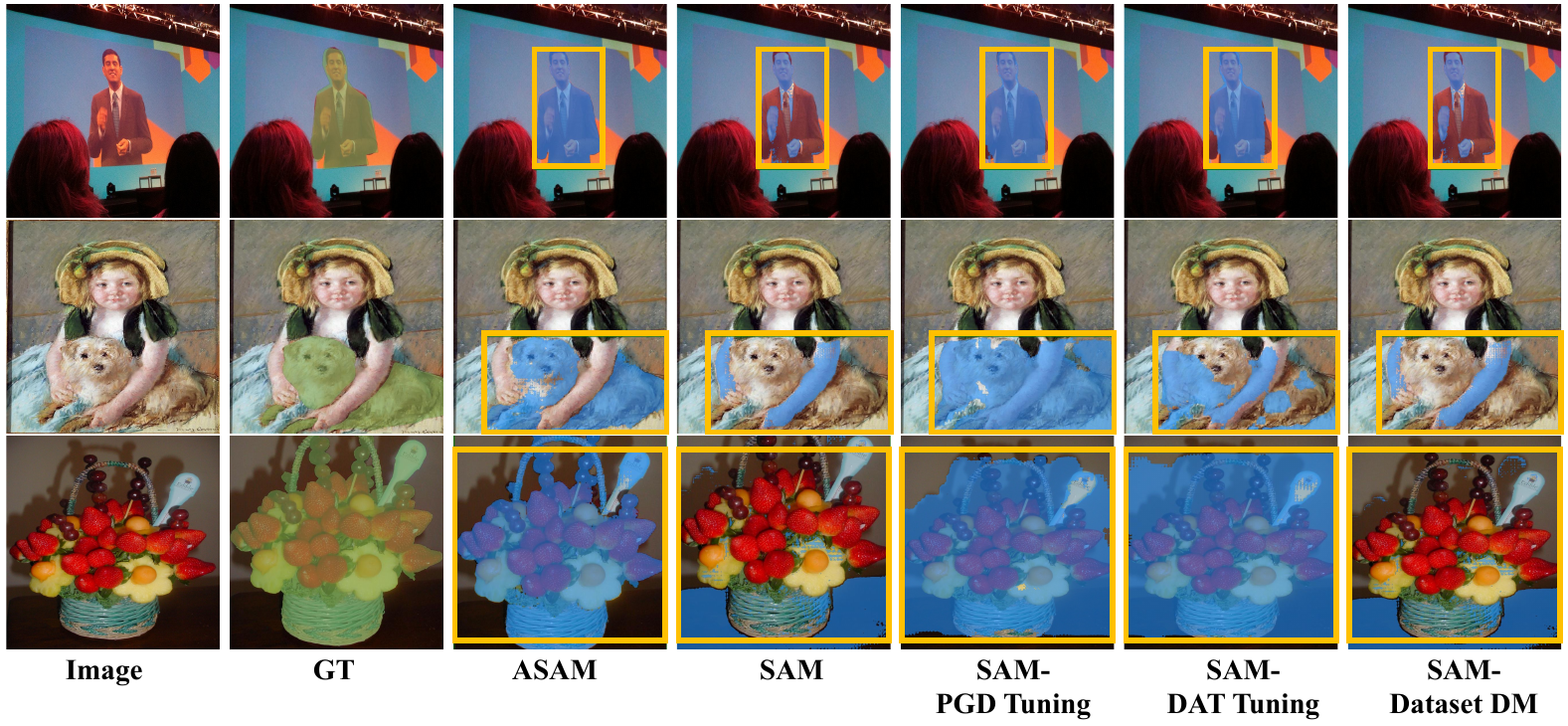}
    \caption{Qualitative comparison of the proposed ASAM and other methods. \textcolor{dullyellow}{Yellow} boxes represent the box prompts.}
    \label{viscomparsion}
    \vspace{-0.5cm}
\end{figure*}

\subsection{Quantitative and Qualitative Comparison}
To thoroughly evaluate the effectiveness of our proposed ASAM, we compare it with four different approaches: the original SAM, SAM fine-tuned with PGD Tuning~\cite{DBLP:conf/icml/RiceWK20}, SAM fine-tuned with DAT Tuning~\cite{DBLP:conf/nips/MaoCDZQYLZ022}, and SAM fine-tuned with new data generated through DatasetDM~\cite{DBLP:journals/corr/abs-2308-06160}. As shown in Table. \ref{totaltable1}, ASAM clearly outperforms the other tuning methods. ASAM, compared to the original SAM, achieves performance improvements across all 14 test datasets, with an average performance increase of \textbf{\textit{1.3}} mIoU. This consistent enhancement across a diverse range of datasets underscores the robustness and effectiveness of our approach, demonstrating its capacity to significantly boost the model's capabilities in various contexts. A key reason for this superiority is that SAM has already been trained on a large-scale dataset. Therefore, simply adding noise perturbations to some samples or generating new samples for tuning SAM does not introduce a significantly different data distribution to SAM. In fact, re-tuning might disrupt the originally well-trained parameters of SAM. Different from existing methods like PGD and DAT, our adversarial samples are reconstructed from a well-optimized, low-dimensional manifold guided by the gradients of SAM. This approach allows us to more effectively address the shortcomings in SAM's original training. It provides a refined input that is better aligned with SAM's learning paradigm, enabling it to generalize more effectively to new or challenging scenarios. From a visual comparison in Fig. \ref{viscomparsion}, it is evident that our proposed ASAM enhances the performance on samples where the original SAM fell short.

\begin{table}[]
\centering
\caption{Ablation studies of main components in ASAM.}
\scalebox{0.9}{
\begin{tabular}{@{}ccc|c@{}}
\toprule
\begin{tabular}[c]{@{}c@{}}Latent \\ Projection\end{tabular} &
  \begin{tabular}[c]{@{}c@{}}Latent \\ Optimization\end{tabular} &
  \begin{tabular}[c]{@{}c@{}}Controllable \\ Generation\end{tabular} &
  mIoU \\ \midrule
$\checkmark$ &              &              & 59.1 \\
$\checkmark$ & $\checkmark$ &              & 54.3 \\
$\checkmark$ &              & $\checkmark$ & 69.3 \\
$\checkmark$ & $\checkmark$ & $\checkmark$ & 77.6 \\ \bottomrule
\end{tabular}}
\label{maincompont}
\vspace{-0.5cm}
\end{table}

\subsection{Ablation Studies}
Herein, we conduct ablation studies on the 14 datasets mentioned above to indicate the effectiveness of ASAM.

\noindent \textbf{Main components.} As shown in Table. \ref{maincompont}, if we solely rely on Latent Projection (Sec. \ref{step1.1}) without employing Latent Optimization (Sec. \ref{step1.2}), performance diminishes as it lacks the guidance from SAM's gradient. This approach misses out on the crucial step of refining the latent representation based on the model's feedback, which is essential for aligning the projection with the model's learned patterns and intricacies. Furthermore, if we use only Latent Projection followed by reconstruction with ControlNet but still omit Latent Optimization, performance again falls short. This combination, while slightly more sophisticated, still fails to leverage the model-specific insights that Latent Optimization provides, thus not fully capitalizing on the potential improvements in the projection process. Finally, when Latent Optimization is combined with ControlNet, we achieve the best segmentation result.

\begin{figure*}
    \centering
    \includegraphics[width=0.75\linewidth]{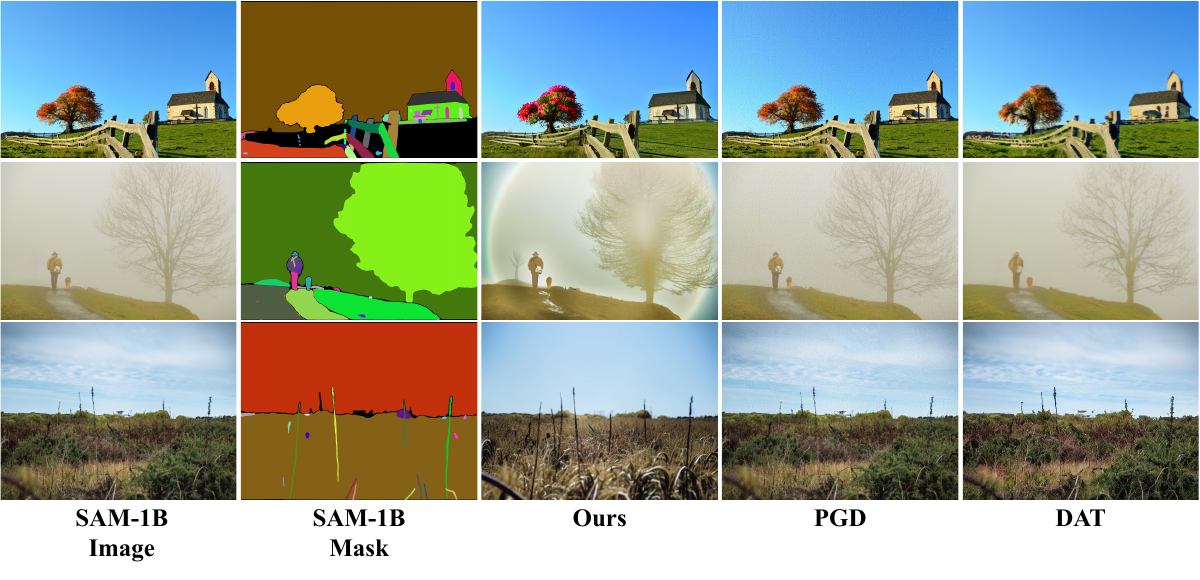}
    \caption{Adversarial examples comparison of ASAM and other attack methods.}
    \label{fig:enter-labeladv}
    \vspace{-0.5cm}
\end{figure*}

\begin{table}[]
\centering
\caption{Image quality assessment.}
\scalebox{0.78}{
\begin{tabular}{@{}c|cccc@{}}
\toprule
Method      & NIMA-AVA$\uparrow$ & HyperIQA $\uparrow$ & MUSIQ -AVA $\uparrow$ & TReS $\uparrow$ \\ \midrule
 SA-1B & 5.18               & 0.72                & 4.07                  & 78.46           \\
Inversion   & 5.19               & 0.72                & 4.11                  & 78.59           \\ \midrule
PGD         & 5.04               & 0.69                & 3.85                  & 76.17           \\
DAT         & 4.76               & 0.63                & 3.83                  & 66.13           \\
Ours        & 5.20               & 0.71                & 4.12                  & 76.73           \\ \bottomrule
\end{tabular}}
\label{qualasse}
\vspace{-0.5cm}
\end{table}

\noindent \textbf{Adversarial Samples Visualization.} 
To validate the utility of the adversarial samples produced in this study for the fine-tuning of SAM, we adopt a quantitative approach to image quality assessment, in line with previous research~\cite{DBLP:conf/cvpr/ShamsabadiSC20,DBLP:conf/nips/YuanZGCS22}. Specifically, we employ non-reference perceptual image quality metrics for this purpose. The metrics selected include NIMA \cite{DBLP:journals/tip/TalebiM18}, HyperIQA \cite{DBLP:conf/cvpr/SuYZZGSZ20}, MUSIQ \cite{DBLP:conf/iccv/KeWWMY21}, and TReS \cite{DBLP:conf/wacv/GolestanehDK22}. Both NIMA-AVA and MUSIQ-AVA have been trained on the AVA dataset \cite{DBLP:conf/cvpr/MurrayMP12}, utilizing the PyIQA framework \cite{IQA-PyTorch}. As depicted in Table. \ref{qualasse}, the inversion images produced in our work maintain comparable image quality to their clean counterparts. Notably, ASAM outperforms other methods in terms of image quality assessment. We further illustrate this with adversarial samples showcased in Fig. \ref{fig:enter-labeladv}. It's important to highlight that the perturbations introduced via ASAM are designed to be natural, in contrast to the more artificial alterations typical of other techniques, such as DAT or PGD tuning methods. This approach to generating natural perturbations aims to create authentically challenging examples akin to those encountered in real-world scenarios, thereby potentially improving the model's generalization capabilities.

\noindent \textbf{Framework Transferability.}
To further assess the transferability of our ASAM framework, we conduct experiments on another large vision foundation model, EfficientSAM (ESAM)~\cite{DBLP:journals/corr/abs-2312-00863}, which is the novel large vision foundation model proposed by Meta in CVPR2024. The results in Table. \ref{esam} corroborate the framework's capability to significantly boost ESAM's performance as well. These findings validate our framework's efficacy across different large models, paving the way for boosting the capabilities of large vision foundation models.

\begin{table}[]
\centering
\caption{ESAM vs AESAM on vit-tiny backbone.}
\scalebox{0.8}{
\begin{tabular}{@{}c|ccccc@{}}
\toprule
Method &
  Ade20k &
  VOC2012 &
  Cityscapes &
  COCO2017 &
  \multicolumn{1}{l}{LVIS} \\ \midrule
ESAM &
  75.0 &
  80.5 &
  54.5 &
  78.4 &
  80.0 \\
AESAM &
  {\color[HTML]{333333} \textbf{76.2}} &
  {\color[HTML]{333333} \textbf{80.8}} &
  {\color[HTML]{333333} \textbf{55.3}} &
  {\color[HTML]{333333} \textbf{79.6}} &
  {\color[HTML]{333333} \textbf{81.3}} \\ \bottomrule
\end{tabular}}
\label{esam}
\vspace{-0.5cm}
\end{table}

\section{Discussion \& Future work}
Although we have demonstrated the effectiveness of our method through extensive empirical experiments, it seems that in addition to the direct inspiration from NLP research, the theoretical underpinnings specific to our method remain an area for further exploration. Fortunately, we have found some existing theoretical work that, although not directly applicable to our task, can provide some theoretical evidence. Specifically, we find that our approach in ASAM aligns with the theoretical framework proposed by Wong and Kolter~\cite{DBLP:conf/iclr/0001K21}, which emphasizes bridging the gap between real-world perturbations and adversarial defenses. This paper underlines the value of learning perturbation sets directly from data, mirroring our method of using the Stable diffusion model to generate natural adversarial examples. Furthermore, the use of Conditional Variational Autoencoders (CVAEs) for perturbation learning in the paper supports our methodology of manipulating latent space representations. These theoretical insights reinforce the effectiveness of using generative models to create adversarial examples that are not just challenging for the model but also reflect real-world complexities and variations.  Although this paper cannot serve as direct theoretical proof for our work, this theoretical backing complements our empirical findings, highlighting the effectiveness of using realistic adversarial examples for enhancing SAM's performance in different real-world scenarios. 

This connection, however, is just the beginning of a broader theoretical exploration. Our future work aims to delve deeper into the theoretical aspects of adversarial fine-tuning, specifically in the context of foundation models. We plan to investigate and formalize the principles underlying the efficacy of our method, which could potentially lead to a more generalized theory for enhancing model performance with adversarial examples in the field of computer vision. By establishing a solid theoretical framework, we can further legitimize the use of such techniques and possibly uncover new avenues for improving foundation models' capabilities in diverse real-world applications.


\section{Conclusion}
ASAM, introduced in this study, represents a significant advancement in the SAM through the innovative use of adversarial tuning. Employing a stable diffusion model to augment a segment of the SA-1B dataset, we generated natural, photorealistic adversarial images, leading to substantial improvements in SAM's segmentation capabilities across various tasks. This method, inspired by adversarial training techniques in NLP, maintains the original architecture and zero-shot strengths of SAM while enhancing its performance. Our findings demonstrate that ASAM not only sets new benchmarks in segmentation tasks but also contributes to the broader application and understanding of adversarial examples in the field of computer vision, offering a novel and effective approach to boosting the capabilities of large vision foundation  models.

\small
\bibliographystyle{ieeenat_fullname}
\bibliography{main}

\end{document}